\documentclass[]{bytedance}
\usepackage[toc,page,header]{appendix}

\usepackage{minitoc}
\usepackage{amsfonts}
\usepackage{amssymb}
\usepackage{tabularx}
\usepackage{listings}
\usepackage{xcolor}
\usepackage{cancel}

\usepackage{tabulary,multirow,xspace}
\usepackage{fixmath,mathtools,nicefrac,mmstyle}
\usepackage{subcaption}
\usepackage{caption}
\usepackage{wrapfig} 
\usepackage[misc]{ifsym} 
\usepackage{colortbl}

\usepackage{wrapfig}
\usepackage{multicol}
\usepackage[most]{tcolorbox}
\usepackage{pifont}
\usepackage{cleveref}
\definecolor{codegreen}{rgb}{0,0.6,0}
\definecolor{codegray}{rgb}{0.5,0.5,0.5}
\definecolor{codepurple}{rgb}{0.58,0,0.82}
\definecolor{backcolour}{rgb}{0.95,0.95,0.92}
\definecolor{boxblue}{RGB}{57,89,163}
\definecolor{boxbluebg}{RGB}{230,237,250} 

\lstdefinestyle{mystyle}{
    backgroundcolor=\color{backcolour},   
    commentstyle=\color{codegreen},
    keywordstyle=\color{magenta},
    numberstyle=\tiny\color{codegray},
    stringstyle=\color{codepurple},
    basicstyle=\ttfamily\footnotesize,
    breakatwhitespace=false,         
    breaklines=true,                 
    captionpos=b,                    
    keepspaces=true,                 
    numbers=none,                    
    numbersep=5pt,                  
    showspaces=false,                
    showstringspaces=false,
    showtabs=false,                  
    tabsize=2
}
\definecolor{mygray1}{gray}{.95}
\definecolor{mygray2}{gray}{.9}
\definecolor{mygray3}{gray}{.95}
\usepackage{pifont}

\newlength\savewidth
\newcolumntype{x}[1]{>{\centering\arraybackslash}p{#1pt}}

\newcommand{\app}{\raise.17ex\hbox{$\scriptstyle\sim$}}

\usepackage{xcolor}
\usepackage{graphicx}
\usepackage{amssymb}
\usepackage{pifont}
\usepackage{floatrow}
\usepackage{amsmath} 
\usepackage{float}
\usepackage{wrapfig}
\usepackage{multirow}
\usepackage{tcolorbox}
\tcbuselibrary{breakable, skins, raster}
\usepackage{listings}
\usepackage{listings}

\definecolor{commentgreen}{rgb}{0.1, 0.4, 0.1}
\definecolor{keywordblue}{rgb}{0.1, 0.1, 0.7}
\definecolor{stringred}{rgb}{0.7, 0.1, 0.1}

\lstdefinestyle{mystyle}{
    commentstyle=\color{commentgreen},
    keywordstyle=\color{keywordblue},   
    stringstyle=\color{stringred},
    basicstyle=\ttfamily\scriptsize, 
    breaklines=true,
    keepspaces=true,
    showstringspaces=false,
    frame=none,                     
    language=Python, 
}

\newcommand{\name}{LibraGen}
\title{\name{}: Playing a Balance Game in Subject-Driven Video Generation}

\author[1,\star]{Jiahao Zhu}
\author[1,\star]{Shanshan Lao}
\author[1,\star]{Lijie Liu}
\author[1,\star]{Gen Li}
\author[1,\star]{Tianhao Qi}
\author[1,\star]{Wei Han}
\author[1,\star\dagger]{Bingchuan Li}
\author[1]{FangFang Liu}
\author[1]{\\Zhuowei Chen}
\author[1]{Tianxiang Ma}
\author[1]{Qian He}
\author[\S]{Yi Zhou}
\author[2,\S]{Xiaohua Xie}

\affiliation[1]{ByteDance}
\affiliation[2]{Pazhou Laboratory (Huangpu), China}
\contribution[\star]{Equal contribution}
\contribution[\dagger]{Project lead}
\contribution[\S]{Corresponding author}

\abstract{
With the advancement of video generation foundation models (VGFMs), customized generation, particularly subject-to-video (S2V), has attracted growing attention. However, a key challenge lies in balancing the intrinsic priors of a VGFM, such as motion coherence, visual aesthetics, and prompt alignment, with its newly derived S2V capability. Existing methods often neglect this balance by enhancing one aspect at the expense of others. To address this, we propose LibraGen, a novel framework that views extending foundation models for S2V generation as a balance game between intrinsic VGFM strengths and S2V capability. Specifically, guided by the core philosophy of ``Raising the Fulcrum, Tuning to Balance,'' we identify data quality as the fulcrum and advocate a quality-over-quantity approach. We construct a hybrid pipeline that combines automated and manual data filtering to improve overall data quality. To further harmonize the VGFM’s native capabilities with its S2V extension, we introduce a Tune-to-Balance post-training paradigm. During supervised fine-tuning, both cross-pair and in-pair data are incorporated, and model merging is employed to achieve an effective trade-off. Subsequently, two tailored direct preference optimization (DPO) pipelines, namely Consis-DPO and Real-Fake DPO, are designed and merged to consolidate this balance. During inference, we introduce a time-dependent dynamic classifier-free guidance scheme to enable flexible and fine-grained control. Experimental results demonstrate that LibraGen outperforms both open-source and commercial S2V models using only ten-thousand-scale training data.
}

\date{\today}

\checkdata[Project Page]{\url{https://phantom-video.github.io/LibraGen/}}

\begin{document}
\maketitle

\section{Introduction}
\begin{figure}[t]
\centering
\includegraphics[width=0.98\textwidth]{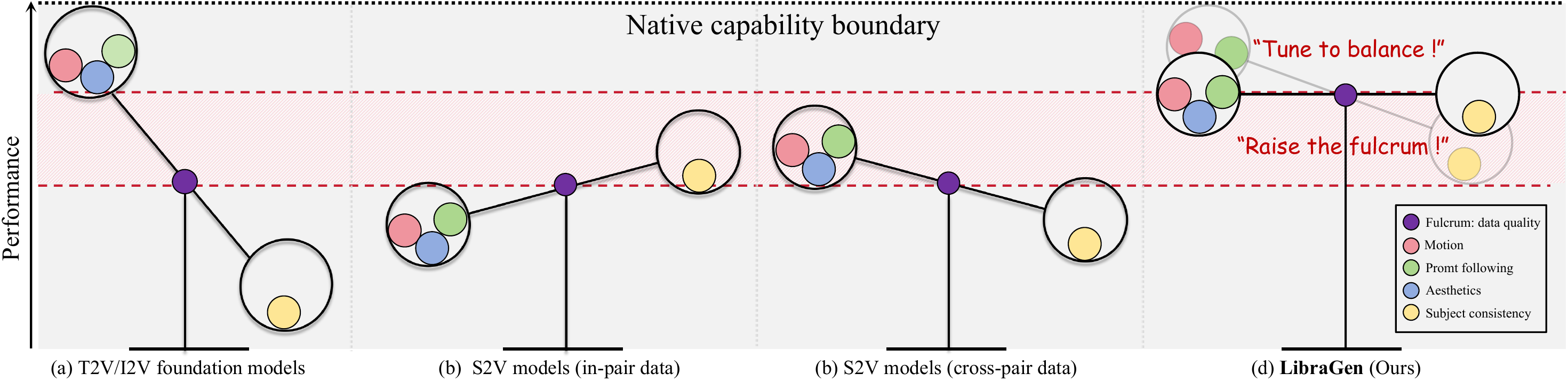}
\caption{\textbf{A balance game in S2V generation}. (a) T2V/I2V foundation models lack task-specific training data and thus exhibit poor S2V performance. (b) Previous S2V methods trained solely on in-pair data or (c) solely on cross-pair data often overlook the inherent balance trade-off. (d) LibraGen frames S2V generation as a balance game, achieving superior and well-balanced S2V performance.}
\label{fig:balancegame}
\end{figure}
\begin{figure}[t]
\centering
\includegraphics[width=0.95\textwidth]{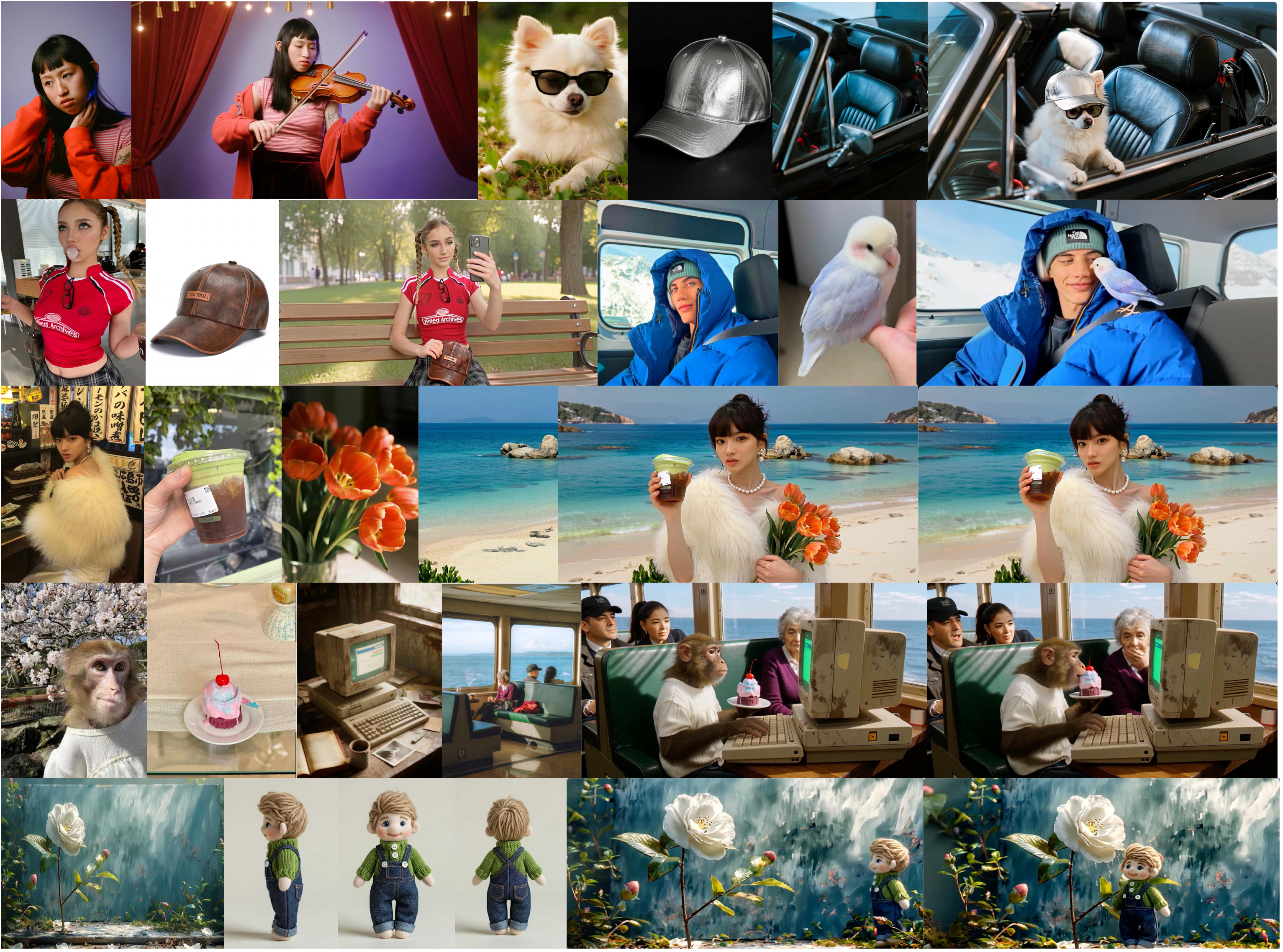}
\caption{We present LibraGen, a novel training paradigm that extends VGFMs to support both single-subject and multi-subject driven video generation.}
\label{fig:visualexamples}
\end{figure}
Recently, numerous open-source \cite{opensora,kong2024hunyuanvideo,yang2024cogvideox,wan2025wan} and commercial \cite{hailuo,kling,Veo3.1,vidu,gao2025seedance} video generation foundation models (VGFMs) have emerged, accelerating intelligent video content creation. These models mainly target two tasks: text-to-video (T2V) and image-to-video (I2V). T2V models generate text-aligned videos, yet prompt-only control struggles to meet complex, multi-modal, fine-grained user demands. I2V models turn static images into dynamic content, but the fixed initial frame constraint severely limits creative flexibility. To enable more customized control, commercial products (e.g., Vidu \cite{vidu}, Kling \cite{kling}, and Sora2\cite{sora2}) have extended their foundation models to the subject-to-video (S2V) task, which generates text-aligned, subject-consistent videos using textual prompts and reference images.

Extending a text-to-video (T2V) or image-to-video (I2V) foundation model to subject-to-video (S2V) generation represents a typical case of task-incremental learning, which inherently necessitates careful balancing between retaining prior task knowledge and acquiring new capabilities \cite{van2022three,wang2024comprehensive}. Consequently, the central challenge in S2V generation is to faithfully preserve the foundation model’s intrinsic advantages, such as superior motion quality, appealing visual aesthetics, and strong prompt adherence, while simultaneously achieving robust subject-consistent video generation. As demonstrated in \cref{fig:balancegame}(a), off-the-shelf T2V/I2V foundation models exhibit markedly poor zero-shot performance on S2V tasks due to the complete absence of tailored training data. To overcome this limitation, a widely established training paradigm has emerged: fine-tuning these foundation models on carefully constructed \texttt{<prompt, video, reference>} triplets. Several studies \cite{chen2025multi,liang2025movie,deng2025cinema,huang2025conceptmaster,xue2025stand,zhou2025scaling} form such triplets by directly extracting the reference subject from frames within the target video clip itself (commonly referred to as in-pair data); however, this seemingly straightforward approach introduces severe reference content leakage during training, frequently resulting in undesirable copy-and-paste artifacts, which in turn cause the generated videos to suffer from degraded visual aesthetics, suppressed motion dynamics, and weakened prompt alignment, as illustrated in \cref{fig:balancegame}(b). More recent methods \cite{liu2025phantom,chen2025humo,deng2025magref,zhang2025kaleido} advocate the use of cross-pair data for fine-tuning, sourcing reference subjects from entirely separate video clips; while this strategy effectively eliminates the copy-and-paste issue and substantially improves prompt responsiveness, it significantly weakens the intrinsic correlation between the reference image and the target video content, often leading to compromised subject consistency, as shown in \cref{fig:balancegame}(c). Furthermore, because most current S2V models are trained predominantly on large-scale, automatically curated (and thus noisy) datasets, ensuring consistently high human-aligned quality remains challenging, ultimately constraining overall S2V performance to sub-optimal levels.

To address these issues, we propose \textbf{LibraGen}, a novel framework that extends T2V/I2V foundation models to support S2V generation by formulating the problem as a principled balance game. The framework is built upon a new training paradigm, termed ``Raising the Fulcrum, Tuning to Balance,'' as shown in \Cref{fig:balancegame}(d), which comprises three key aspects:

\textbf{Data Curation}. Recent VGFMs exhibit strong T2V and I2V capabilities, greatly lowering the barrier to S2V learning and thus necessitating a paradigm shift from data quantity to data quality. We argue that data quality acts as a critical balancing fulcrum, and its careful refinement can significantly boost overall S2V performance. To this end, we pursue three key directions: (1) Data refinement—we distill a million-scale raw dataset into a thousand-scale, human-aligned high-quality subset through a systematic automatic–manual hybrid pipeline encompassing video collection, precise subject extraction, and accurate captioning; (2) Data trait analysis—we strategically leverage the complementary strengths of in-pair and cross-pair data by thoroughly considering their inherent characteristics; (3) Data labeling—we introduce a hierarchical tagging system that enables fine-grained control over the training data distribution.

\textbf{Model Training}. We propose a Tune-to-Balance post-training paradigm that harmonizes the foundation model’s inherent strengths with its derived S2V capability. During supervised fine-tuning (SFT), models trained on in-pair and cross-pair data exhibit complementary strengths and weaknesses in subject consistency and foundation model capabilities. We adopt a weighted model merging strategy to make a trade-off. We further design two direct preference optimization (DPO) pipelines, termed \textbf{Consis-DPO} and \textbf{Real-Fake DPO}, and merge them to consolidate this balance, where the former enhances subject consistency and the latter restores foundation model performance.

\textbf{Model Inference}. We observe that early denoising steps primarily capture text-guided structural layouts and motion patterns, while later steps increasingly incorporate fine-grained visual details from the reference. Based on this, we propose a time-dependent dynamic classifier-free guidance (CFG) strategy that adaptively adjusts the relative strengths of text and reference conditioning throughout the denoising process, enabling precise and flexible control over the balance between prompt adherence and reference fidelity.

The contributions of this paper are summarized as follows:
\begin{itemize}
\item \textbf{Concept}. This paper identifies the fundamental challenge in S2V generation as an intrinsic trade-off between preserving the native VGFM capabilities and ensuring robust subject-consistent video generation. 
\item \textbf{Technology}. We propose LibraGen, a novel training paradigm that regards S2V generation as a balance game. It introduces a quality-over-quantity data curation strategy and a Tune-to-Balance post-training framework.
\item \textbf{Performance}. As shown in \Cref{fig:visualexamples}, while trained on thousand-scale data, LibraGen delivers impressive single- and multi-subject driven video generation, outperforming both commercial and open-source S2V models.
\end{itemize}
\section{Related Works}
\subsection{Video Generation Foundation Model}
Recent advances in diffusion models have fueled rapid progress in video generation. Early video generation models \cite{ho2022video,singer2022make,chen2023videocrafter1,zeng2024make} typically adopted 2D U-Net backbones, integrating temporal modeling components (\emph{e.g.}, attention mechanisms or convolutional layers) prior to fine-tuning on specialized video datasets. However, their generative performance is inherently constrained by architectural design and model scale. With the continuous expansion of computational resources, scaling laws \cite{kaplan2020scaling,yin2025towards} have been empirically validated in the video generation domain, spurring the development of large-scale video generation models based on the DiT architecture \cite{peebles2023dit,esser2024sd3}; the videos synthesized by these models demonstrate remarkably superior visual fidelity. On the open-source front, representative models include Wan \cite{wan2025wan}, HunyuanVideo \cite{kong2024hunyuanvideo}, and CogVideoX \cite{yang2024cogvideox}, whereas closed-source counterparts, such as Veo 3.1 \cite{Veo3.1}, Kling 2.5 \cite{kling}, Hailuo 2.3 \cite{hailuo}, Vidu \cite{vidu}, and Seedance 1.0/1.5 \cite{gao2025seedance,chen2025seedance}, have attained commercial-grade performance levels. Fundamentally, foundation models for video generation primarily focus on T2V and I2V tasks, with their design objectives anchored in three core pillars: motion coherence, visual quality, and prompt alignment. Leveraging their scalability, these models can be adapted to a wide range of downstream tasks via customized post-training strategies.

\subsection{Subject-to-Video Generation}
The objective of S2V generation is to generate subject-consistent and prompt-aligned videos from given reference images and textual instructions. This can be achieved by fine-tuning state-of-the-art video generation foundation models on \texttt{<prompt, video, reference>} triplets. Some studies \cite{chen2025multi,liang2025movie,deng2025cinema,huang2025conceptmaster,xue2025stand,zhou2025scaling} construct such triplets (\emph{a.k.a.} in-pair triplets) by extracting reference subjects from frames within target video clips. However, this strategy often leads to copy-and-paste artifacts. Although some methods \cite{chen2025multi} alleviate this issue through data augmentation or prompt engineering, they do not fundamentally address the problem of unnatural integration of reference subjects, such as lighting inconsistencies between the subjects and the background in generated videos. To overcome these issues, subsequent studies \cite{liu2025phantom, chen2025humo, deng2025magref, zhang2025kaleido} propose fine-tuning with cross-pair triplets, where reference subjects are extracted from different video clips. By weakening the direct correlation between target videos and paired reference images, this strategy effectively mitigates copy-and-paste artifacts.
\section{Model Design}
\subsection{Lightweight Subject Injection}
\begin{figure*}[!t]
\centering
\includegraphics[width=0.98\textwidth,height=6cm]{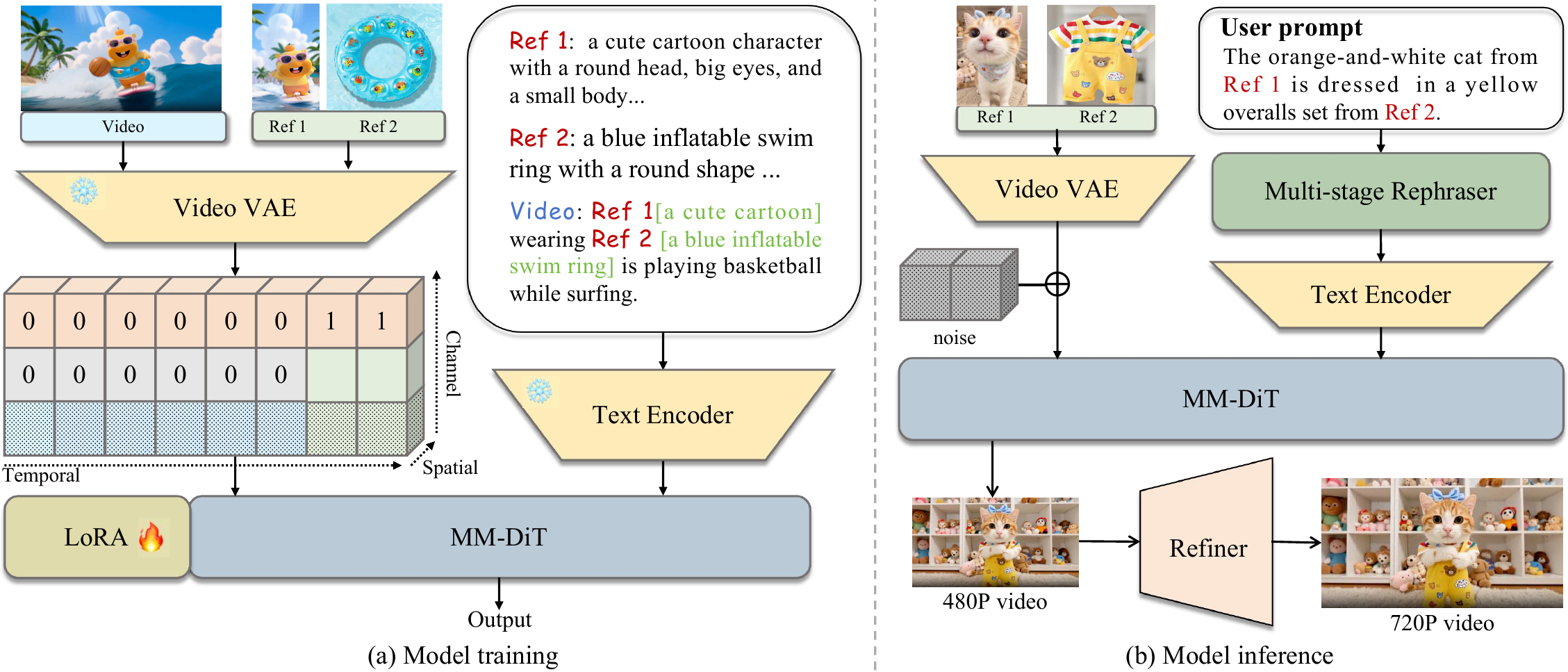}
\caption{\textbf{Training and inference pipelines of LibraGen}. During training, reference-related embeddings are concatenated with video frame embeddings along the temporal dimension, accompanied by dedicated flags. During inference, the user prompt is processed by a multi-stage prompt rephraser to better align with the training captions. The base MM-DiT produces 480P outputs, which can be further refined to 720P.}
\label{fig:modedesign}
\end{figure*}
We build LibraGen upon a Multi-Modal Diffusion Transformer (MM-DiT) \cite{gao2025seedance}, which consists of low- and super-resolution DiTs with interleaved spatial and temporal attention blocks. To enable S2V generation with minimal impact on the foundation model, we adopt a lightweight subject injection framework that requires neither additional vision encoders \cite{yuan2025identity,chen2025multi,liu2025phantom} nor extra preprocessing steps such as subject segmentation \cite{chen2025multi,deng2025magref,zhang2025kaleido}. As illustrated in \Cref{fig:modedesign}, during training, reference images and video frames are first encoded by the same VAE, yielding embeddings $\mathbf{z}^{\mathrm{r}_i}\in \mathbb{R}^{n\times c\times h\times w}$ and $\mathbf{z}^{\mathrm{v}}\in\mathbb{R}^{f\times c\times h\times w}$, respectively. The reference and frame embeddings are then concatenated along the temporal dimension:
\begin{equation}
\mathbf{z}_t^{\mathrm{noised}}=\mathrm{Concat}([\mathbf{z}_t^{\mathrm{v}},\mathbf{z}_t^{\mathrm{r}_i}],\mathrm{dim}=0)\in\mathbb{R}^{(f+n)\times c\times h\times w},
\end{equation}
where $\mathbf{z}_t^*=(1-t)\mathbf{z}^*+t\boldsymbol{\epsilon}$ within $\boldsymbol{\epsilon}\sim\mathcal{N}(\mathbf{0},\mathbf{I})$. Reference conditions are constructed by padding the reference embeddings with zeros:
\begin{equation}
\mathbf{z}^{\mathrm{cond}}=\mathrm{Pad}([\mathbf{z}^{\mathrm{r}_i},\mathbf{0}],\mathrm{dim}=0)\in\mathbb{R}^{(f+n)\times c\times h\times w},
\end{equation}
We then concatenate $\mathbf{z}^{\mathrm{cond}}$ with $\mathbf{z}_t^{\mathrm{noised}}$ along the channel dimension, together with dedicated binary flags $\mathbf{f}$ to indicate reference embeddings:
\begin{equation}
\mathbf{z}_t^{\mathrm{input}}=\mathrm{Concat}([\mathbf{z}_t^{\mathrm{noised}},\mathbf{z}^{\mathrm{cond}},\mathbf{f}],\mathrm{dim}=1)\in\mathbb{R}^{(f+n)\times(2c+1)\times h\times w}.
\end{equation}

\subsection{Multi-stage Prompt Rephraser} During inference, the user prompt, as shown in \Cref{fig:modedesign}, is often misaligned with the training captions (detailed in \Cref{sec:data_curation}), which may degrade generation quality. To bridge this gap, we propose a multi-stage prompt rephrasing strategy. In the first stage, we use Qwen 3 \cite{bai2025qwen2} to generate fine-grained descriptions for each reference subject. In the second stage, the VLM extracts distinguishable coarse-level descriptions of the reference subjects and merges them with the user prompt. In the third stage, the processed prompt is appropriately elaborated by adding relevant details about subject motion and scene context.
\section{Model Training}
\begin{figure*}[!t]
\centering
\includegraphics[width=0.98\textwidth]{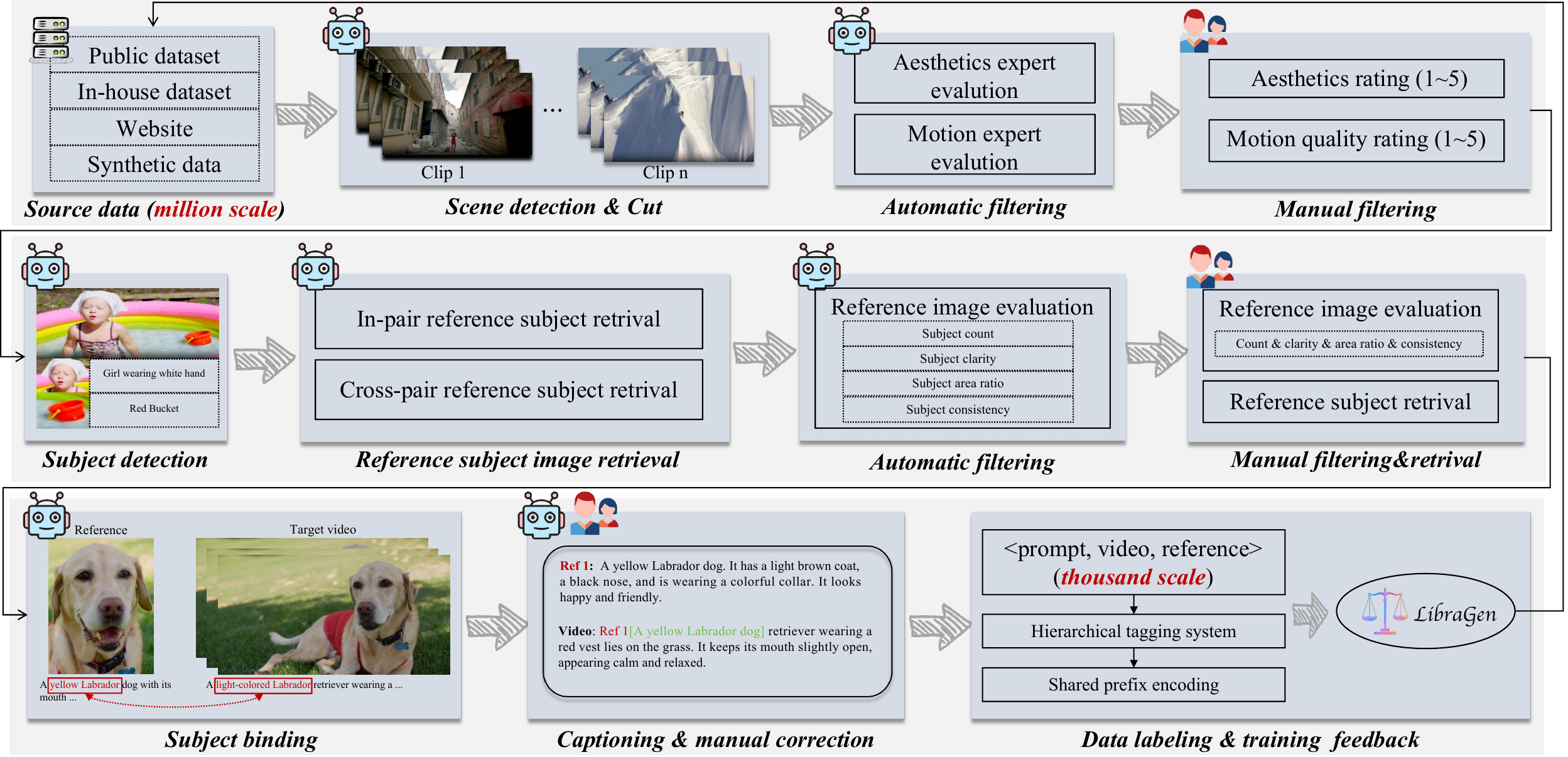}
\caption{\textbf{Human-aligned data curation pipeline}. The automatic–manual hybrid data curation pipeline follows a quality-over-quantity strategy and consists of four stages: video collection, reference subject extraction, data captioning, and data labeling with dynamic updates.}
\label{fig:pipeline}
\end{figure*}

We frame the extension of T2V/I2V foundation models to S2V as task-incremental learning, which requires a balance between pre-trained knowledge and novel task acquisition \cite{van2022three,wang2024comprehensive}. We propose a quality-over-quantity approach, featuring a data curation pipeline that generates thousand-scale human-aligned training samples (\Cref{fig:pipeline}). Furthermore, we introduce the Tune-to-Balance post-training paradigm to effectively harmonize existing and new model capabilities.
\subsection{Human-aligned Data curation}
\label{sec:data_curation}

\noindent\textbf{Video Collection}. Our video data originates from diverse sources, including Panda70M~\cite{chen2024panda}, an in-house collection of 3 million video clips, online platforms, and synthesized content. Long videos are segmented using AutoShot~\cite{zhu2023autoshot} and PySceneDetect \cite{PySceneDetect} into 2–11 second clips. We employ a dual-stage filtering process: first, clips are scored by motion and aesthetics experts \cite{liu2025improving}, retaining the top 5$\%$. Second, we conduct manual evaluations for motion coherence, amplitude, and aesthetics on a 1–5 scale, keeping only those with an average score of 4 or higher. This pipeline ensures strong motion dynamics and high visual quality across our dataset.

\noindent\textbf{Reference Subject Extraction}. We adopt the S2V detection pipeline~\cite{chen2025phantom-data} to extract bounding boxes of primary subjects from keyframes. These subjects are further cropped and matched against a large-scale retrieval bank~\cite{chen2025phantom-data} to construct \texttt{<video, reference>} pairs. For cross-pair data, we retrieve the top 15 most similar samples per query subject, excluding the source video, and randomly select one as the reference. For in-pair data, subjects detected within the same video clip serve as references. To ensure quality, we filter ambiguous subjects, enforce subject consistency for cross-pair data, and constrain subject scales to prevent background leakage in in-pair data. This process combines automated VLM-based filtering with manual verification and data supplementation.

\noindent\textbf{Data Captioning}. We caption each \texttt{<video, reference>} pair in the form of \texttt{<ref$_i$ text, video text>}. We first caption each reference subject and then perform subject binding using the video description and metadata \texttt{<subject, bounding box, reference>} provided by Stage 2. For in-pair data, where reference images are directly drawn from the target video, the subject name in the video description (highlighted in red \cref{fig:modedesign}) is replaced with the refined reference description (highlighted in green). For cross-pair data, we identify the corresponding reference for each subject using the metadata and perform the replacement accordingly. Finally, manual verification is conducted to ensure accurate data captioning.

\noindent\textbf{Data Labeling and Dynamic Updating}. Each \texttt{<prompt, video, reference>} triplet is annotated using a hierarchical tagging system. In this work, we employ a five-level architecture. Categories at each level are assigned unique identifiers, and each triplet is represented via shared-prefix encoding. This labeling scheme enables us to efficiently analyze statistics and dynamically adjust the training data distribution based on model feedback.
\subsection{Supervised Fine-tuning}
\noindent\textbf{Low-resolution}. To preserve the inherent capabilities of the foundation model to the greatest extent, we adopt Low-Rank Adaptation (LoRA) \cite{hu2022lora} for fine-tuning. Recent studies \cite{liu2025phantom,deng2025magref} have demonstrated that fine-tuning exclusively on in-pair data gives rise to severe copy-paste artifacts and degraded prompt-following performance. In contrast, training solely on cross-pair data mitigates these issues but compromises subject consistency. To address this dilemma, we propose fine-tuning two distinct models on in-pair and cross-pair data separately, and then linearly interpolating their respective LoRA weights to achieve an adjustable trade-off.

\noindent\textbf{Super-resolution}. To enable subject-consistent super-resolution, reference subjects are incorporated to guide the super-resolution process. We construct quadruple data \texttt{<prompt, 480P video, 720P video, reference>} for super-resolution SFT. Specifically, high-resolution videos are sampled from the dataset constructed in \cref{sec:data_curation}, and their low-resolution counterparts are sythesized via downsampling combined with slight noise-induced blurring to simulate real-world quality degradation. Since the super-resolution network shares a similar architecture with the low-resolution model \cite{gao2025seedance}, we initialize it by merging the LoRA obtained in low-resolution SFT phase.  Using the constructed quadruple data, we perform fine-tuning with an exponential moving average.
\subsection{Tailored DPO} 
\begin{figure*}[!t]
\centering
\includegraphics[width=0.95\textwidth]{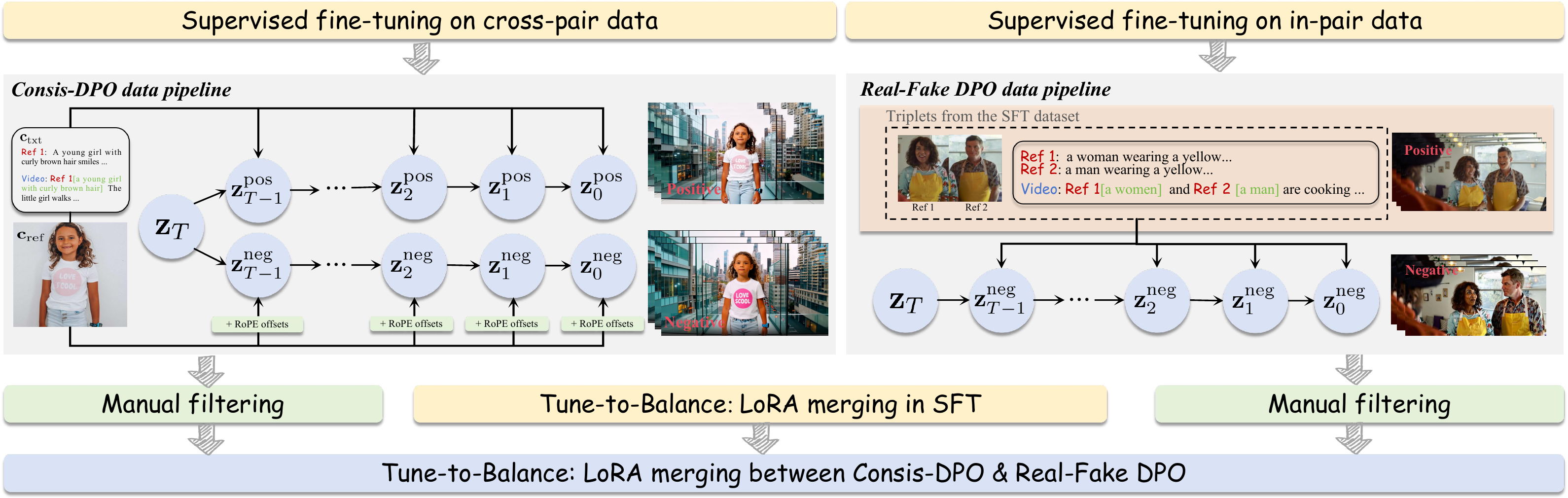}
\caption{\textbf{Tune-to-Balance post-training paradigm}.}
\label{fig:post_training}
\end{figure*}

\noindent\textbf{Consis-DPO}. To further enhance subject consistency, we propose a subject-consistency-enhanced DPO strategy. The key challenge is to construct positive–negative pairs that differ significantly in subject consistency while minimally impacting other dimensions, such as motion and visual quality. To address this, we design a dedicated pipeline that synthesizes such pairs by manipulating the RoPE offsets of reference images during model inference. Specifically, the positive sample is generated by the SFT model trained on cross-pair data since cross-pair data yield good visual and motion effects. The corresponding negative sample is produced using an identical inference configuration, including the same initial noise, sampling steps, and conditions with the only modification being increased RoPE offsets applied to reference-related embeddings in the temporal attention blocks, as illustrated in \Cref{fig:post_training}. We generate a large number of such positive–negative candidate pairs and then carefully curate a high-quality preference dataset at the thousand-sample scale. Only pairs in which the positive sample demonstrates clearly superior subject consistency are retained.

\noindent\textbf{Real-Fake DPO}. While Consis-DPO effectively improves subject consistency, it tends to slightly degrade performance on other evaluation dimensions in practice, particularly motion dynamics. This degradation arises mainly from the low quality of positive samples generated by the SFT model and DPO’s inherent preference hacking issue \cite{fisch2024robust, gupta2025mitigating}. To address these problems, we propose a Real–Fake pair data curation pipeline. Specifically, positive samples are directly selected from training triplets during the SFT stage, which already exhibit strong visual quality and motion dynamics. The corresponding negative samples are generated by the model fine-tuned on in-pair data, using identical prompts and reference images as the positive counterparts. This design is motivated by the observation that the in-pair fine-tuned model tends to produce copy-and-paste artifacts, a desirable property for constructing effective negative samples. We then construct a large candidate set of positive–negative pairs and carefully curate a high-quality dataset at the thousand scale by retaining only pairs where negative samples maintain strong subject consistency but suffer from poor visual and motion quality.

Building upon the LoRA-merged SFT model, we also adopt LoRA to enable efficient DPO training, with the training objective defined as
\begin{equation}
    \begin{aligned}
        \mathcal{L}_{\text{DPO}}(\theta) = -\mathbb{E} \left[ \log \sigma \left( \beta \log \frac{p_\theta(\mathbf{z}^{\mathrm{pos}}|\mathbf{c}_\mathrm{txt},\mathbf{c}_\mathrm{ref})}{p_{\theta_{\mathrm{ref}}}(\mathbf{z}^{\mathrm{pos}}|\mathbf{c}_\mathrm{txt},\mathbf{c}_\mathrm{ref})} - \beta \log \frac{p_\theta(\mathbf{z}^{\mathrm{neg}}|\mathbf{c}_\mathrm{txt}, \mathbf{c}_\mathrm{ref})}{p_{\theta_{\mathrm{ref}}}(\mathbf{z}^{\mathrm{neg}}|\mathbf{c}_\mathrm{txt},\mathbf{c}_\mathrm{ref})} \right) \right].
    \end{aligned}
\end{equation} 
$\theta_{\mathrm{ref}}$ is initialized from the low-resolution SFT model and updated via an exponential moving average of $\theta$ after each iteration. In addition, we employ LoRA to train two models separately on paired data generated by Consis-DPO and Real–Fake DPO. Finally, we merge the LoRA weights to achieve a desirable trade-off.
\subsection{Time-dependent dynamic CFG}
\label{sec:dynamic_cfg}
During inference, we follow methods \cite{liu2025phantom,chen2025humo} to adopt CFG with separate
guidance scales for each modality:
\begin{equation}
    \begin{aligned}
        &\mathbf{v}_{\theta}(\mathbf{z}_t,\mathbf{c}_{\mathrm{txt}},\mathbf{c}_{\mathrm{ref}})= \mathbf{v}_{\theta}(\mathbf{z}_t,\oslash,\oslash) + \omega_1\left(\mathbf{v}_{\theta}(\mathbf{z}_t,\mathbf{c}_{\mathrm{ref}},\oslash) - \mathbf{v}_{\theta}(\mathbf{z}_t,\oslash,\oslash)\right) \\
        & + \omega_2\left(\mathbf{v}_{\theta}(\mathbf{z}_t,\mathbf{c}_{\mathrm{ref}}, \mathbf{c}_{\mathrm{txt}}) - \mathbf{v}_\theta(\mathbf{z}_t,\mathbf{c}_{\mathrm{ref}},\oslash)\right),
    \end{aligned}
\end{equation}
where $\omega_1$ and $\omega_2$ control the guidance scales of the reference subjects $\mathbf{c}_{\mathrm{ref}}$ and text prompts $\mathbf{c}_{\mathrm{txt}}$, respectively. 

During the early stages of denoising, DiT predominantly captures structural information, including spatial object layouts and temporal motion trajectories, which are primarily guided by text prompts. In contrast, later stages emphasize visual details such as textures and colors, which are more strongly influenced by reference images \cite{lv2025dual}. Motivated by this observation, we introduce a time-dependent dynamic CFG strategy, in which $\omega_1$ progressively increases over time $t$ while $\omega_2$ correspondingly decreases. This strategy enables flexible and fine-grained control over both reference images and textual prompts without causing unexpected visual degradation.

\section{Experiment}
\subsection{Setup}
\noindent \textbf{Implementation Details}. 
In the low-resolution SFT phase, we train on 9,000 samples (5,500 cross-pair and 3,500 in-pair). We increase the reference image drop ratio for in-pair training (to mitigate copy-paste artifacts) and the text drop ratio for cross-pair training (to boost subject consistency), with early stopping at 8,000 and 5,000 iterations respectively. For super-resolution SFT, 4,000 low-resolution samples are selected to build quadruple data, followed by 5,000 more fine-tuning iterations. In the DPO phase, Consis-DPO and Real-Fake pipelines generate 2,000 and 1,600 positive-negative pairs, respectively, with 4,000 iterations per training run.

The SFT objective is based on Rectified Flow \cite{liu2022flow}, adopting the noise sampling strategy from \cite{esser2024scaling}. During training and inference, each window in temporal DiT blocks only perceives a partial reference subject, which may impair subject consistency. We thus introduce the all-gather strategy to enable full reference subject capture.

\noindent \textbf{Evaluation Dataset and Baselines}. 
To evaluate LibraGen, we select Kling O1 \cite{kling-omni-2025} and Vidu Q1 \cite{vidu} as commercial baselines, and Phantom \cite{liu2025phantom} and MAGREF \cite{deng2025magref} as open-source baselines. Our evaluation utilizes an in-house benchmark of 200 diverse test cases encompassing both single- and multi-subject generation. The dataset is distributed across 1, 2, 3, and 4 reference images with proportions of $26\%$, $50\%$, $14\%$, and $10\%$, respectively. To test real-world robustness, the benchmark includes uncropped reference images where subjects may occupy only a small fraction of the frame, spanning categories such as animated IPs, humans, and consumer products.

\noindent \textbf{Evaluation Metrics.}
To evaluate performance, we employ Motion Smoothness (MS) \cite{huang2024vbench} and Motion Quality (MQ) \cite{liu2025improving} to quantify temporal fluidity and general motion fidelity. Visual appeal and perceptual integrity are assessed using the Aesthetic Score (AES) and Image Quality Assessment (IQA) from the VBench \cite{huang2024vbench}. To capture human preferences more accurately, we incorporate the Visual Quality (VQ) reward model \cite{liu2025improving}. Furthermore, semantic consistency between text prompts and generated videos is measured via the Text Alignment (TA) reward model \cite{liu2025improving}. To assess subject consistency, we calculate the GSB Ratio, defined as $(G-B)/(G+S+B)$, to determine the relative win rate against baseline models. In this metric, $G$, $S$, and $B$ denote instances where our model is superior to, comparable to, or inferior to the reference, respectively.
\begin{table*}[!t]
\centering
\small
\setlength{\tabcolsep}{8pt}

\begin{tabular}{lcccccc}\toprule
\multirow{2}{*}{Method} & \multicolumn{2}{c}{Motion Quality} & \multicolumn{3}{c}{Visual Quality} & \multicolumn{1}{c}{Text Align.}\\
\cmidrule(lr){2-3} \cmidrule(lr){4-6} \cmidrule(lr){7-7}

 & MS $\uparrow$ & MQ $\uparrow$ & AES $\uparrow$& IQA $\uparrow$& VQ $\uparrow$ & TA $\uparrow$\\ \midrule

Vidu Q1 \cite{vidu}             & \underline{0.5373} & \underline{0.9924} & \underline{0.6491} & \underline{71.82}& \underline{2.795} & 3.315\\
Kling O1 \cite{kling-omni-2025} & 0.4965 & 0.9865 & 0.6479 & \textbf{72.84} & \textbf{2.797} & 3.567 \\
MAGREF \cite{deng2025magref}    & 0.3830 & 0.9853 & 0.6356 & 70.36 & 2.354 & 1.784 \\
Phantom \cite{liu2025phantom}   & 0.3844 & 0.9873 & 0.6410& 69.35 & 2.373 & 1.998 \\ 
\midrule
LibraGen (ours)               & \textbf{0.5380} & \textbf{0.9930} & \textbf{0.6496}& 71.60& \underline{2.795} & \textbf{3.594} \\ \bottomrule
\end{tabular}
\caption{\textbf{Quantitative evaluation of motion, visual quality, and text alignment} using text prompts and reference images as inputs. The best results are shown in bold, and the underline indicates the second-best performance.}
\label{tab:s2v_result}
\end{table*}
\begin{table*}[!t]
\centering
\setlength{\tabcolsep}{10pt}
\begin{tabular}{lcccc}
\bottomrule
Baseline& \multicolumn{1}{c}{1} & 2 & 3 & 4 \\ \midrule
Vidu Q1 \cite{vidu}             & 0.154& 0.140& 0.143& 0.100\\
Kling O1 \cite{kling-omni-2025} & 0.077& 0.080& 0.071& 0.100\\
MAGREF \cite{deng2025magref}& 0.423& 0.620& 0.429& 0.700\\ 
Phantom \cite{liu2025phantom}   & 0.308& 0.300& 0.286& 0.500\\ \bottomrule
\end{tabular}
\caption{\textbf{GSB ratio results relative to different S2V baselines}. We compare subject consistency in single- and multi-subject scenarios (2, 3, and 4 reference images). Higher values indicate better subject consistency performance of LibraGen.}
\label{tab:GSB_ratio}
\end{table*}
\begin{figure}[!t]
    \centering
    \includegraphics[width=0.95\textwidth]{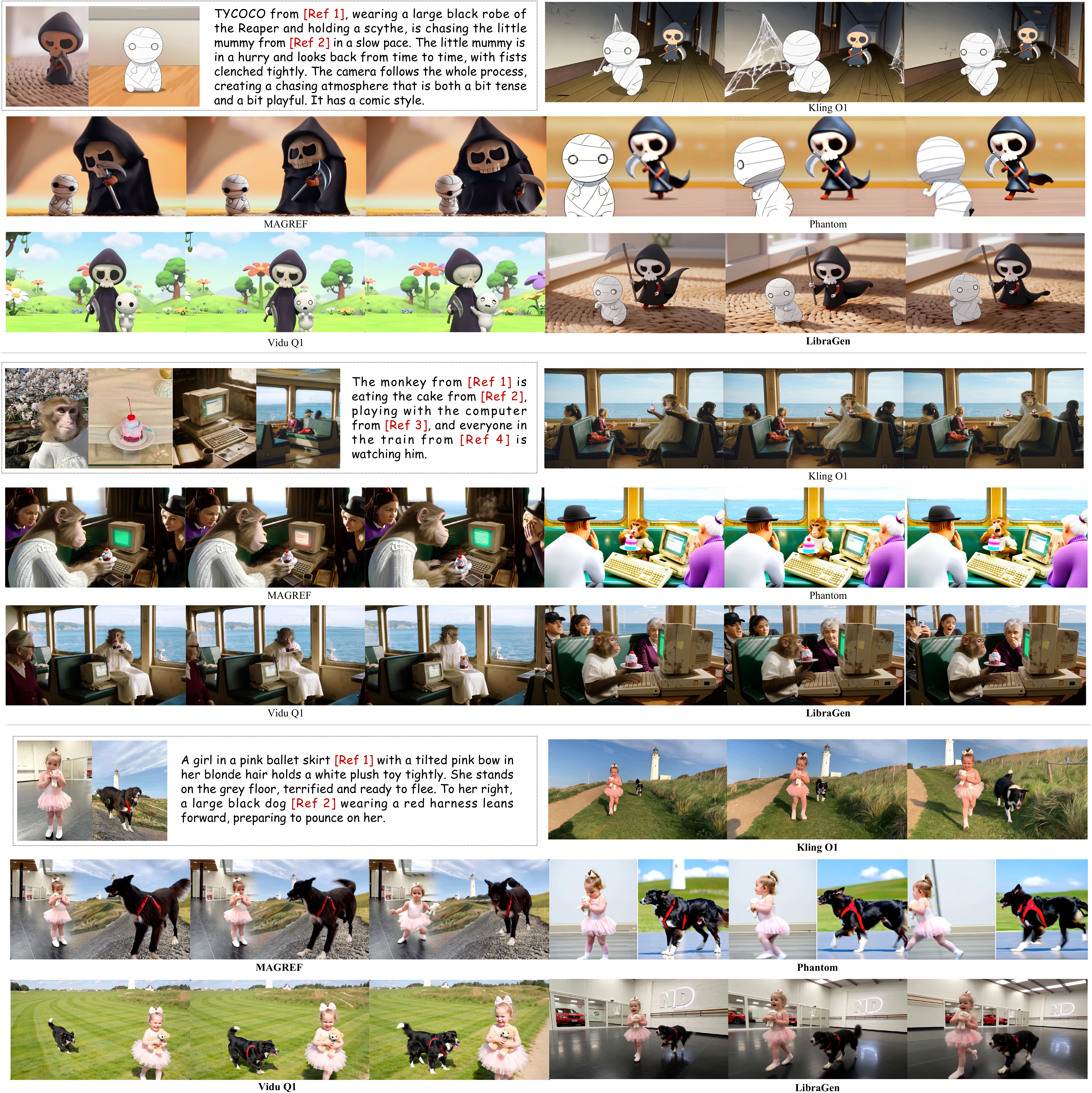}
    \caption{\textbf{Visual Comparison with state-of-the-art S2V models}. First, middle, and last frames are sampled for each method for visualization. Please zoom in for visual details.}
    \label{fig:visual_comp}
    \end{figure}
\subsection{Comparison with Other Methods}
As reported in Table 1, LibraGen achieves superior performance across multiple dimensions. In terms of motion dynamics, it leads in both Motion Smoothness and Motion Quality, underscoring its ability to generate fluid and physically consistent movements. Regarding visual perception, LibraGen obtains the best AES and remains highly competitive in IQA and VQ, performing on par with leading models like Kling O1 and Vidu Q1. Furthermore, LibraGen achieves state-of-the-art performance in prompt following, surpassing other S2V models and ensuring that the generated videos remain semantically faithful to the input prompts.

The core strength of LibraGen in maintaining subject identity is further substantiated by the GSB ratio results reported in \Cref{tab:GSB_ratio}. Across all experimental settings, ranging from single-subject to intricate multi-subject configurations, LibraGen consistently achieves positive GSB ratios against all baselines, demonstrating a significantly higher win rate regarding subject fidelity. Notably, our approach exhibits marked superiority over MAGREF, with GSB ratios peaking at 0.700. Notably, empirical observations indicate that Phantom outperforms MAGREF within our evaluation framework; this discrepancy is largely attributable to MAGREF’s condition injection mechanism, which involves concatenating all reference images into a single frame, a constraint requiring high spatial occupancy by subjects for effective encoder capture. Consequently, when presented with test cases involving small-scale subjects in the reference images, MAGREF inevitably suffers from performance degradation. These findings highlight LibraGen’s robustness in preserving fine-grained subject details and ensuring identity stability, effectively addressing the persistent challenges of identity blurring and semantic drifting inherent in contemporary subject-driven generation frameworks. We further provide a visual comparison in \Cref{fig:visual_comp} to intuitively showcase the effectiveness of our approach. Notably, LibraGen maintains robust subject consistency when handling multi-subject driving tasks.
\begin{figure}[!t]
\centering
\includegraphics[width=0.90\textwidth]{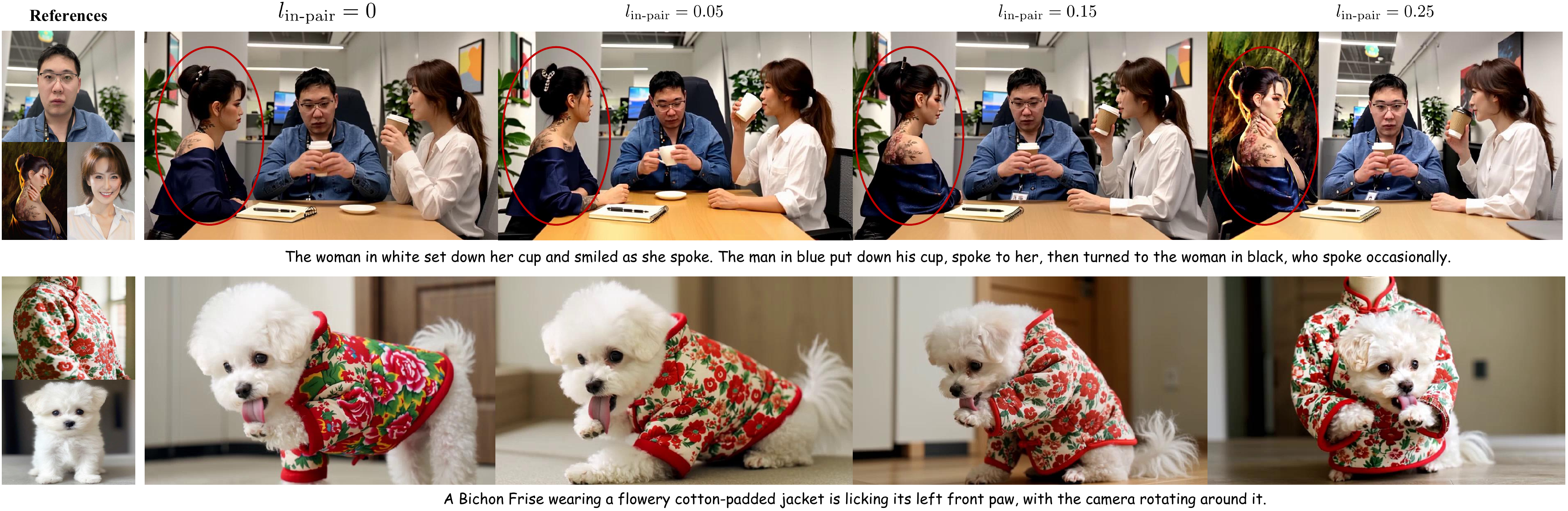}
\caption{\textbf{Merging the LoRA module trained on in-pair data in SFT phase}. As the merging coefficient increases, subject consistency gradually improves, but the copy-paste phenomenon becomes increasingly severe. }
\label{fig:ablation1}
\end{figure}
\begin{figure}[!t]
\centering
\includegraphics[width=0.85\textwidth]{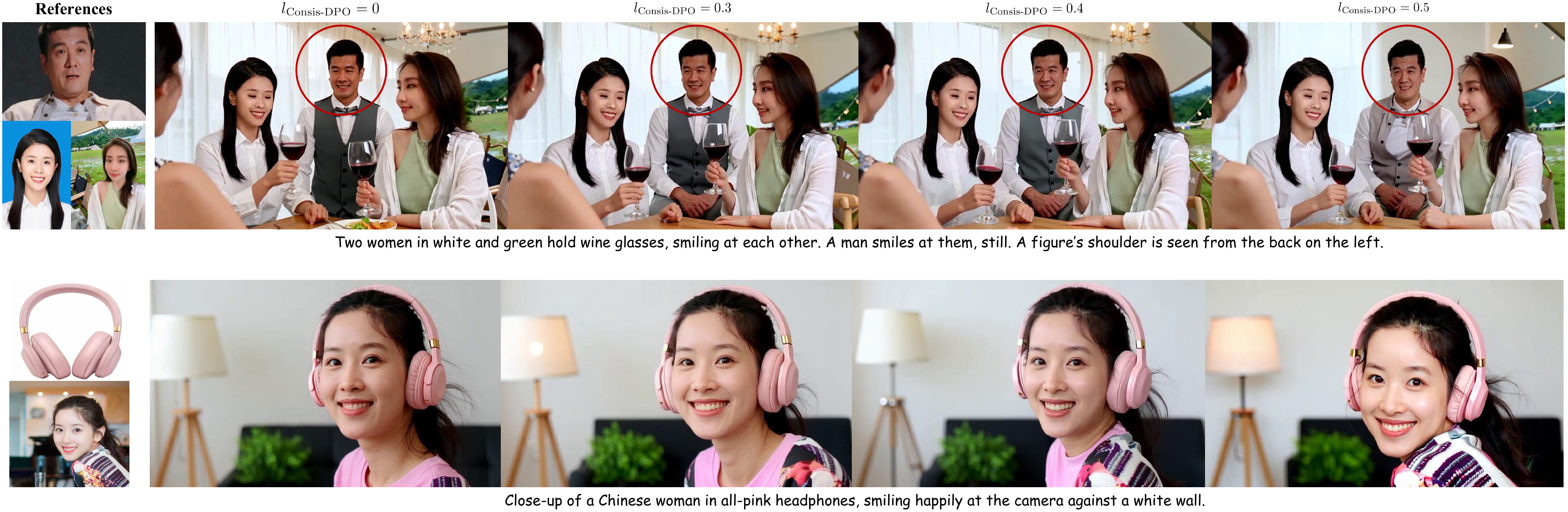}
\caption{
\textbf{Merging the LoRA module trained on data generated through the Consis-DPO pipeline}. Consis-DPO can effectively improve human identity consistency without introducing the copy-paste phenomenon.}
\label{fig:ablation2}
\end{figure}
\begin{table*}[!t]
\centering
\small
\begin{tabular}{lccccccc} 
\toprule
\multirow{2}{*}{Strategy} & \multicolumn{2}{c}{Motion Quality} & \multicolumn{3}{c}{Visual Quality} & Text Align. & Subject Consist. \\
\cmidrule(lr){2-3} \cmidrule(lr){4-6} \cmidrule(lr){7-7} \cmidrule(lr){8-8}
& MS $\uparrow$ & MQ $\uparrow$ & AES $\uparrow$ & IQA $\uparrow$ & VQ $\uparrow$ & TA $\uparrow$ & GSB ratio $\uparrow$ \\ 
\midrule
baseline                       & 0.9912   & 0.5350  & 0.6420   & 71.87 & 2.794   & 3.504  & - \\
+ $l_{\text{in-pair}}=0.15$          & 0.9894   & 0.5218   & 0.6306  & 71.45 & 2.708  & 3.466  & 0.190 \\
+ $l_{\text{Consis-DPO}}=0.5$    & 0.9907   & 0.5239   & 0.6334  & 71.10 & 2.714  & 3.490    & 0.286  \\
+ $l_{\text{Real-fake}}=0.1$        & 0.9913   &  0.5342  & 0.6394  & 71.42  & 2.756   & 3.524  & 0.025 \\
+ dynamic CFG       & 0.9930   &  0.5380  & 0.6496  & 71.60  & 2.795   & 3.594  & 0.020 \\
\bottomrule
\end{tabular}
\caption{\textbf{Analysis of training and inference strategies}. Note that ``$+$'' indicates a cumulative addition to the baseline. GSB ratios are benchmarked against the previous setup, where the metric is calculated regardless of the reference subject count.}
\label{tab:ablation_training}
\end{table*}
\subsection{Training and Inference Strategy Analysis}
\textbf{Tune-to-balance in SFT}. In the SFT phase, the model is first trained on cross-pair data. However, as evidenced by the results with $l_{\text{in-pair}}=0$ in \Cref{fig:ablation1}, this approach leads to suboptimal subject consistency. Cross-pair training diminishes the alignment between reference subjects and target videos, causing the model to overly prioritize text prompts at the expense of visual cues. To mitigate this issue, we incorporate a LoRA module fine-tuned on in-pair data. As shown in \Cref{fig:ablation1}, increasing the merging coefficient $l_{\text{in-pair}}$ improves subject fidelity; however, excessively high values induce copy-and-paste artifacts due to overfitting to the reference pose. We therefore adopt $l_{\text{in-pair}}=0.15$ to strike an effective balance. According to the ablation results in \Cref{tab:ablation_training}, merging the in-pair LoRA at this weight yields only a minor compromise in motion smoothness and visual quality.

\textbf{Tune-to-balance in RL}. Simply merging the LoRA module trained on in-pair data still yields suboptimal subject consistency, as evidenced by the results in \Cref{fig:ablation2}.  \Cref{fig:ablation2} shows that human identity similarity improves progressively as the merging coefficient $l_{\text{Consis-DPO}}$ increases, demonstrating the effectiveness of Consis-DPO. Notably, Consis-DPO does not further compromise visual quality, motion quality, or text alignment, as reported in \Cref{tab:ablation_training}. This resilience is primarily attributed to our use of cross-pair data for constructing positive samples; this strategy mitigates ``model hacking'' (\emph{e.g.}, background replication) and encourages the model to focus specifically on identity consistency. Finally, to bolster the motion and visual quality of the generated videos, we integrate a Real-Fake LoRA model with a coefficient $l_{\text{real-fake}}=0.1$. As indicated in \Cref{tab:ablation_training}, this merging significantly enhances both motion and visual quality, while subject consistency and text alignment remain robust without degradation.

\textbf{Tune-to-balance in Inference}. The aforementioned results were generated using a static CFG strategy with $\omega_1=\omega_2=3.5$. In this section, we evaluate a dynamic strategy where $\omega_2$ linearly decreases from 5 to 1, while $\omega_1$ increases from 1 to 4. As shown in \Cref{tab:ablation_training}, this dynamic CFG schedule enhances text alignment without degrading other metrics, demonstrating its effectiveness. However, this strategy inevitably increases inference latency, as it requires two additional velocity function computations compared to the standard case where $\omega_1 = \omega_2$.

\section{Conclusion}
We present LibraGen, a principled framework for S2V generation that unifies the strengths of T2V/I2V foundation models with robust subject-consistent video generation. We show that data quality, rather than quantity, is a key factor in improving overall S2V performance. Furthermore, we propose a novel post-training paradigm guided by the philosophy of ``Raising the Fulcrum, Tuning to Balance.'' Experiments demonstrate that LibraGen achieves state-of-the-art performance, surpassing competitive open-source and commercial methods. We believe that LibraGen offers a promising roadmap for adapting off-the-shelf T2V/I2V foundation models to S2V tasks without requiring full fine-tuning or million-scale training data. Moreover, the design principles of LibraGen are extensible and can be readily applied to emerging audio-video generation models.

\clearpage

\bibliographystyle{plainnat}
\bibliography{main}
\end{document}